\documentclass[11pt,a4paper]{article}
\usepackage[hyperref]{acl2020}
\usepackage{times}
\usepackage{latexsym}

\usepackage{microtype}

\usepackage{graphicx}
\usepackage{colortbl}
\definecolor{lightgray}{gray}{0.9}
\usepackage{amsmath}
\usepackage{amsthm}
\usepackage{amssymb}
\usepackage{multirow}
\usepackage{enumitem}
\theoremstyle{plain}
\newtheorem{thm}{Theorem}
\theoremstyle{definition}
\newtheorem{defn}[thm]{Def.} 

\aclfinalcopy 
\def\aclpaperid{864} 

\title{In Layman’s Terms: Semi-Open Relation Extraction from Scientific Texts}

\author{Ruben Kruiper$^{*}$, Julian F.V. Vincent, Jessica Chen-Burger, \\ {\bf \large Marc P.Y. Desmulliez, Ioannis Konstas}\\ 
    Heriot-Watt University\\
    Riccarton Campus, EH14 4AS\\
    Edinburgh, United Kingdom\\
    $^{*}$Corresponding author: \texttt{rk22@hw.ac.uk}
    }

\date{}

\begin{document}
\maketitle
\begin{abstract}
Information Extraction (IE) from scientific texts can be used to guide readers to the central information in scientific documents. But narrow IE systems extract only a fraction of the information captured, and Open IE systems do not perform well on the long and complex sentences encountered in scientific texts. In this work we combine the output of both types of systems to achieve Semi-Open Relation Extraction, a new task that we explore in the Biology domain. First, we present the Focused Open Biological Information Extraction (FOBIE) dataset and use FOBIE to train a state-of-the-art narrow scientific IE system to extract trade-off relations and arguments that are central to biology texts. We then run both the narrow IE system and a state-of-the-art Open IE system on a corpus of 10k open-access scientific biological texts. We show that a significant amount (65\%) of erroneous and uninformative Open IE extractions can be filtered using narrow IE extractions. Furthermore, we show that the retained extractions are significantly more often informative to a reader.\footnote{We release FOBIE and code at \url{https://github.com/rubenkruiper/FOBIE}.}
\end{abstract}

 

\section{Introduction}
Identifying the central theme and concepts in scientific texts is a time-consuming task for experts and a hard task for laymen \cite{Alper2004,El-Arini2011,Pain2017}. This problem is even more pronounced in inter-disciplinary fields of study, where experts in a target domain often lack the deeper knowledge of a source domain \cite{Carr2018}. A specific example is biomimetics, an engineering problem-solving process in which one draws on analogous biological solutions \cite{Kruiper2016}. 
A major issue is that engineers (target domain) know little biology (source domain) or characteristics of plants or animals \cite{Vattam2013a}. This domain-mismatch complicates searching for and reasoning over relevant scientific information, rendering biomimetics adventitious and solutions serendipitous \cite{Kruiper2018}.


\begin{figure}[!t]
    \centering
    \includegraphics[width=\linewidth]{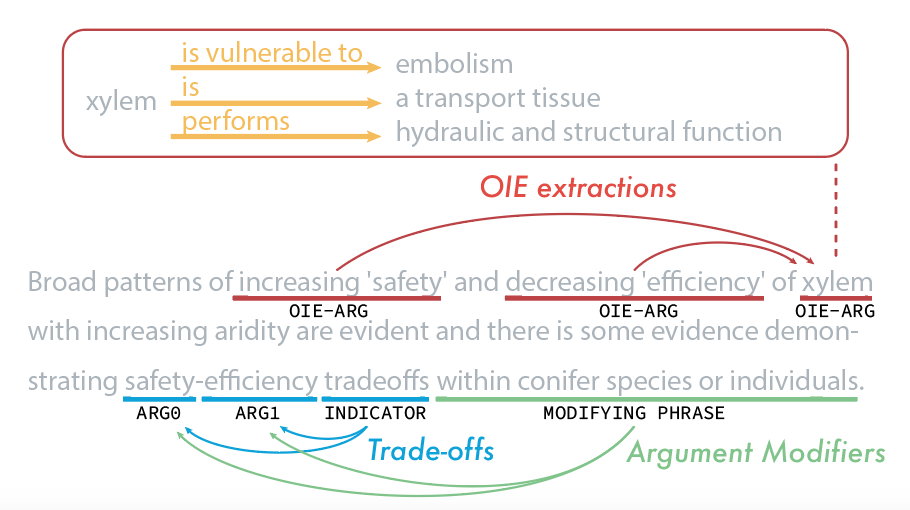}
    \caption{
    Example of Semi-Open Relation Extraction in the Biology domain \cite{Burgess2006}. We first extract the \textsc{Trade-Off} expressed between the central concepts `\textit{safety}' and `\textit{efficiency}' (in blue), that takes place specifically in `\textit{conifer species}' (in green). 
    We further explore the content of a paper by investigating the results of an Open Information Extraction (OIE) system (in red). The central concepts captured by a \textsc{Trade-Off} mechanism enable the filtering of many irrelevant OIE extractions, which are found to be error-prone in scientific texts.
    Further OIE extractions found in the document can shed light on the semantic meaning of relevant concepts, e.g., `\textit{xylem}', as depicted at the top of the Figure (in the red box).
    } \label{fig:Figure1} 
\end{figure}

Recently, \textsc{Trade-Off} relations have become of interest to biomimetics \cite{Adriaens2019Evomimetics:2.0} 
because a trade-off defined in technology can be directly used to search for relevant texts in biology \cite{Vincent2016}. \textsc{Trade-Off} relations express a problem space in terms of mutual exclusivity constraints between competing demands. Therefore, trade-offs play a prominent role in evolutionary thinking \cite{Agrawal2010} and are the principal relation under investigation in a significant portion of biology research papers \cite{Garland2014}. The functional demands that are traded off are usually abstract and domain-independent terms, such as `\textit{safety}' and `\textit{efficiency}' in Figure~\ref{fig:Figure1}. A gap remains in quickly comprehending the central information in a text, e.g., the biological mechanisms that are used to manipulate a trade-off.

Information Extraction (IE), and specifically Relation Extraction (RE), can improve the access to central information for downstream tasks \cite{Santos2015,Zeng2014,Jiang2016,Miwa2016,Luan2018a}. However, the focus of current RE systems and datasets is either too \textit{narrow}, i.e., a handful of semantic relations, such as `\textsc{Used-For}' and `\textsc{Synonymy}', or too \textit{broad}, i.e., an \textit{unbounded} number of generic relations extracted from large, heterogeneous corpora \cite{Niklaus2018}, referred to as Open IE (OIE) \cite{Etzioni2005,Banko2007}. Narrow approaches to IE from scientific text \cite{Augenstein2017,Gabor2018,Luan2018a} cover only a fraction of the information captured in a paper -- usually what is within an abstract. It has been shown that scientific texts contain many unique relation types and, therefore, it is not feasible to create separate narrow IE classifiers for these \cite{Groth2018}. On the other hand, OIE systems are primarily developed for the Web and news-wire domain and have been shown to perform poorly on scientific texts. What laymen really need is a bit of both: the accuracy of narrow RE systems to extract central relations from scientific texts \textit{and} the flexibility of an OIE system to capture a much larger fraction of the possible relations expressed in scientific texts.

This work aims to enable rapid comprehension of a large scientific document by identifying a) the central concepts in a text and b) the most significant relations that govern these central concepts. 
To this end, we introduce the task of Semi-Open Relation Extraction (SORE); Figure~\ref{fig:Figure1} illustrates the SORE process.
First, we find 
the central concepts `\textit{safety}' and `\textit{efficiency}' involved in a \textsc{Trade-Off} relation. 
Then, by using the argument concepts of the relation as anchor points, we can explore further concepts and relations, e.g., `\textit{xylem}' in Figure~\ref{fig:Figure1}. Uncovering these relations can elucidate the meaning of unfamiliar concepts to a layperson \cite{Mausam2016}.
The SORE approach is hypothesized to reduce the number of uninformative extractions without limiting RE to a finite set of relations, which could generally benefit IE from scientific articles, e.g., materials discovery \cite{Kononova2019} and drug-gene-mutation interactions \cite{Jia2019}.

To address SORE we create the Focused Open Biological Information Extraction (\textsc{FOBIE}) dataset. \textsc{FOBIE} includes manually-annotated sentences that express explicit trade-offs, or syntactically similar relations, that capture the central concepts in full-text biology papers. We train a span-based RE model used in a strong scientific IE system \cite{Luan2018a} to jointly extract these relation structures. We explore SORE and use the output of our model to filter the output of an OIE system \cite{Saha2018,Saha2017,Pal2016,Christensen2011} on a corpus of biology papers. Qualitative analyses show that the output of a narrow RE model can speed up expert analysis of trade-offs in biological texts, and be used to filter out both erroneous and uninformative OIE extractions.  

\section{Related work} \label{sec:relation_extraction}

\subsection{Open Information Extraction}
OIE systems use a set of handcrafted or learned extraction rules and rely on dependency features to extract open-domain relational tuples from text \cite{Yu2017, Niklaus2018}. As OIE systems rely on syntactic features they require little fine-tuning when applied to different domains and the extraction rules work for a variety of relation types \cite{Mausam2016}. These properties can be especially useful on scientific texts where additional knowledge on unknown concepts can ease the textual comprehension for non-experts. Consider the example OIE extractions for `\textit{xylem}' in the top part of Figure~\ref{fig:Figure1}. 

Existing OIE systems have been shown to perform significantly worse on the longer and more complex sentences found in scientific texts than on Wikipedia texts \cite{Groth2018}. Common issues of OIE systems on Web, News, and Wikipedia texts include the correct identification of the boundaries of an argument, handling latent \textit{n}-ary relations, difficulty handling negations, and generating uninformative extractions \cite{Schneider2017}. Groth \textit{et al.} \shortcite{Groth2018} evaluate the output of two state-of-the-art OIE systems based on correctness, rather than, e.g., the number of missed extractions. They note that the crux of the IE challenge is that extractions reflect the consequence of the sentence. As an example of an uninformative extraction Fader \textit{et al.} \shortcite{Fader2011} note how `\textit{(Faust, made, a deal)}' captures the consequence, but not the critical information of whom Faust made a deal with in the sentence ``\textit{Faust made a deal with the devil.}''. In this work, we explore filtering both incorrect and uninformative OIE extractions from scientific texts using the central concepts that we extract through narrow IE (cf. Section~\ref{sec:uninformative}).


\subsection{Narrow Relation Extraction from scientific text}
Narrow RE entails identifying two or more related entities in a text and classifying the relation that holds between them. 
Early works on the combined task of Named Entity Recognition and labeling of relations between extracted entities used pre-computed dependency features \cite{Liu2013,Chen2015,Lin2016}, word position embeddings \cite{Zeng2014}, or considered only the Shortest Dependency Path between two entities as input \cite{Bunescu2005,Santos2015,Zeng2015}.
Later work aimed to reduce errors propagated by pre-computed dependency features \cite{Nguyen2015}, or by joint modeling of entities and relations \cite{Miwa2016}. Poor performance of these RE systems on scientific texts has led to the development of domain-specific datasets\footnote{\textsc{ScienceIE} SemEval 2017: 500 paragraphs from full-text Computer Science, Material Science, and Physics journal articles, SemEval 2018: 500 abstracts within the domain of Computational Linguistics. \textsc{SciERC}: 500 abstracts from Artificial Intelligence conference and workshop proceedings.}.

The \textsc{ScienceIE} dataset focuses on the extraction of 3 types of key-phrases, rather than Named Entities, and hyponymy and synonymy relations between these \cite{Augenstein2017}.
The SemEval 2018 task 7 dataset focuses on 6 narrow relations between 7 entity types \cite{Gabor2018}. And 
the \textsc{SciERC} dataset 
focuses on 7 relation types, including co-reference, between 6 types of entities \cite{Luan2018a}.
Top systems developed for both SemEval tasks adapt the LSTM-based approach of Miwa \& Bansal \shortcite{Miwa2016}, combined with semi-supervised learning and ensembling \cite{Ammar2018TheExtraction}, as well as pre-trained concept embeddings \cite{Luan2018}.

\begin{table}[]
    \small
    \centering
    \begin{tabular}{c|cccc}
         \#     & FOBIE & \textsc{ScienceIE} & SE'18 & \textsc{SciERC}  \\ \hline
         Arguments      & 5835    & 9946  & 7483 & 8089  \\
         Relations      & 4788   & 672   & 1595 & 4716  \\
         R/doc        & 3.1*   & 1.3   & 3.2  & 9.4  \\
    \end{tabular}
    \caption{Number of arguments, relations and relations per instance for FOBIE, \textsc{ScienceIE}, the SemEval 2018 task 7 dataset and \textsc{SciERC}. R/doc stands for relations per sentence* for FOBIE (and per abstract or paragraph for the other datasets).}
    \label{tab:datasets_comparison}
\end{table}{}

\subsection{BioNLP and BioCreAtIvE}
In the past, several BioNLP and BioCreAtIvE shared tasks were organized that aimed at identifying relations in the biology domain \cite{Hirschman2005, Kim2009, Nedellec2013, Zhou2014}. Many datasets focus primarily on a predefined set of biomedical relations, such as interactions between known proteins, genes, diseases, drugs, and chemicals \cite{Kim2003, Krallinger2017, Cohen2017, Dogan2019}. Examples of more biology-oriented corpora include the \textsc{BB} corpus \cite{Deleger2016} and the \textsc{SeeDev} corpus \cite{Chaix2016}. The \textsc{BB} corpus includes 4 entity types and 2 relation types that revolve around microorganisms of food interest. Besides abstracts and titles, it contains paragraphs and sentences from 20 full-text documents \cite{Bossy2019}.  Similarly, \textsc{SeeDev} consists of 86 paragraphs from 20 full-text articles about seed development in a specific plant, the \textit{Arabidopsis thaliana}. Considering the small size of the dataset, a relatively large number of many entity and relation types are used; 16 types of Named Entities and 21 types of relations. This results in an imbalanced dataset with 7 relations making up less than 1\% of all relations. Furthermore, there is some overlap in source documents for the train/dev/test split \cite{Chaix2016}. 

\noindent
In contrast to the previously described datasets, \textsc{FOBIE} does not classify arguments of relations into specific entity-types. \textsc{FOBIE} contains annotations of key-phrases found in full-text scientific papers, similar to \textsc{ScienceIE}. The key-phrases and relations are annotated in 1,548 relatively long and complex sentences, which were sourced from 1,292 full-text scientific biological texts using a Rule-Base System. Table~\ref{tab:datasets_comparison} provides an overview of the size of \textsc{FOBIE} in comparison to \textsc{ScienceIE}, the SemEval 2018 task 7 dataset and \textsc{SciERC}. Both the \textsc{BB} and \textsc{SeeDev} corpus contain approximately 3,500 relations within a small sub-domain of biology, while \textsc{FOBIE} focuses more generally on the domain of biology. Section~\ref{sec:dataset} describes the collection of \textsc{FOBIE} and dataset statistics in detail.



\section{Dataset description} \label{sec:dataset}
\subsection{Dataset collection} \label{ssec:dataset_creation}
A variety of words are able to indicate a trade-off, e.g., \textit{compromise, optimization, balance, interplay} and \textit{conflict} \cite{Kruiper2018}. We adapt these terms as trigger words in a Rule-Based System (\textsc{RBS}) and run it on 10k open-access papers that were collected from the Journal of Experimental Biology (JEB) and BioMed Central (BMC) journals on `\textit{Biology}', `\textit{Evolutionary Biology}, and `\textit{Systems Biology}'. The selection of journals was made only to the extent that the articles focus on the biological domain. We retained the abstract, introduction, results, discussion and conclusion sections. We used spaCy\footnote{https://spacy.io/} to split the texts into sentences and identify POS tags and dependency structure. The \textsc{FOBIE} dataset contains only sentences that the \textsc{RBS} identified as expressing a \textsc{Trade-off} relation.

\begin{figure*}[ht]
    \includegraphics[width=1\linewidth]{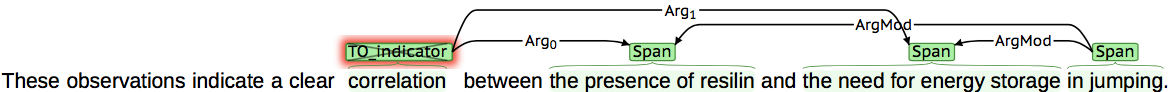}
    \caption{Example of an annotation in BRAT showing a trigger word, `\textit{correlation}', that is related to two arguments, which in turn are related to a single modifier. The trigger word does not indicate a \textsc{Trade-Off} relation, but a positive correlation.}
    \label{fig:modifier_example}
\end{figure*}

\subsection{Annotation} \label{ssec:dataset_annotation}
The initial annotations extracted by the \textsc{RBS} were manually corrected and extended by a biology expert using the BRAT interface \cite{Stenetorp2012}. We define three relation types: \textsc{Trade-off}, \textsc{Argument-Modifier} and \textsc{Not-a-Trade-off}. The latter denotes phrases that are related to a trigger word, but not by a \textsc{Trade-off} relation. These syntactically similar relations provide useful training signal as negative samples. Negative samples are important because possible trigger words can be contiguous, e.g., the phrase `\textit{negative correlation}' denotes a \textsc{Trade-Off} relation, whereas  `\textit{correlation}' by itself does not. As a result, the annotation of training examples is harder, and lexical and syntactic patterns that correctly signify the relation are sparse \cite{Peng2017}. For simplicity's sake, with some abuse of terminology, we refer to all such relations collectively as \textit{trade-offs}.

We found a substantial amount of arguments to be nested or in a non-projective relationship. In Figure~\ref{fig:modifier_example} the prepositional phrase `\textit{in jumping}', conceptually refers to both central concept arguments of the relation, i.e., `\textit{the need for energy storage}' and `\textit{the presence of resilin}'. We adopt the following annotation heuristic: prepositional phrases are treated as modifying phrases when they apply to multiple arguments (as is the case in Figure~\ref{fig:modifier_example}) or can be distinctly separated from the argument, e.g., by punctuation. 

We randomly selected 250 sentences (16.1\%) for re-annotation and quality control by a second domain expert. The inter-annotator agreement Cohen \textit{k} is found to be 0.93. Table~\ref{tab:FOBIE_description} summarizes statistics on FOBIE. The final dataset consists of 1,548 single sentences from 1,292 unique documents, split into 1,248/150/150 train/dev/test. The split is controlled for source document overlap to avoid having identical arguments of relations appearing both during training and testing. \textsc{FOBIE} contains relatively long key-phrases with an average of 3.46 tokens and only 20\% of them consist of a single token. In comparison \textsc{ScienceIE} and \textsc{SciERC} both contain 31\% singleton key-phrases, and the average entity length in \textsc{SciERC} is 2.36. Furthermore, sentences taken from full-text documents are longer than those found in abstracts. The average sentence length in \textsc{SciERC} is 24.31 tokens, while 82.62\% of the sentences in \textsc{FOBIE} are longer than 25 tokens.

\begin{table}[]  
\small
\centering
\begin{tabular}{l|l} 
        \hline
        \# Sentences                    & 1548    \\
        Avg. sent. length               & 37.81   \\
        \% of sents $\geq$ 25 tokens    & 82.62\% \\\hline
        \textbf{Relations}: &\\
        - \textsc{Trade-Off}            & 765     \\
        - \textsc{Not-a-Trade-Off}      & 2502    \\
        - \textsc{Arg-Modifier}         & 1521    \\\hline
        Triggers                        & 1600    \\
        Keyphrases                      & 4235    \\
        Spans                           & 5835    \\
        Unique spans                    & 3643 \\
        Unique triggers                 & 41 \\
        Max triggers/sent               & 2 \\ 
        Max spans/sent                  & 9  \\
        Keyphr. w/ multiple rel's       & 1951 \\
        \# single-word arguments        & 864  \\
        Avg. tokens per argument        & 3.46 \\
\end{tabular}
\caption{Aggregated statistics for FOBIE.}
\label{tab:FOBIE_description}
\end{table}

\begin{figure*}[h]
    \centering
    \includegraphics[width=\textwidth]{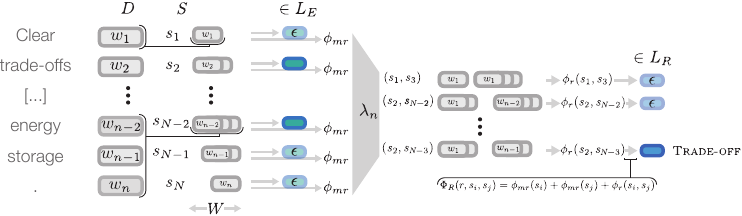}
    \caption{We provide the \textsc{SciIE} system with single sentences $D$ as input. For all possible spans up to width $W$ a span label $\in L_{E}$ is computed and a mention score $\phi_{mr}$. Spans with the lowest mention scores are pruned, with variable beam size $\lambda_{n}$. For combinations of remaining spans a relation label $\in L_{R}$ is predicted. The set of span labels $L_{E}$ and the set of relation labels $L_{R}$ both contain a dummy class $\epsilon$.}
    \label{fig:example_input}
\end{figure*}{}

\section{Narrow IE baseline}
\subsection{Task definition}
Following Peng \textit{et al.} \shortcite{Peng2017} we extract \textit{n}-ary relations by (1) identifying the trigger and (2) extracting the binary relations between this trigger and the arguments -- inspired by Davidsonian semantics. We define key-phrases as spans of consecutive words $ s \in S$, with $S$ all possible spans in a sentence, and relation-types as $r \in R^{d}$, with $d$ the total number of unique relations. Then a binary relation is a triple ${<}governor, relation, dependent{>}$ with $governor$ and $dependent$ elements of $S$. The union of the following binary relations found in a sentence may constitute a non-projective graph:
\begin{defn} \label{def:trade-off}
An explicit trade-off is an instance of a directed relation $t \in T^{o}$, indicated by trigger word $p \in P^{u}$ with $u$ the set of unique trigger words and  $P \subset S$. A trade-off is a binary relation, $t \models o$, with $governor \in P$ and $dependent \in S$. A single trigger word $p$ can be in $n$ multiple relations.
\end{defn}
\begin{defn} \label{def:argmod}
An argument-modifier is a directed binary relation $a \in A^{m}$, where we omit the classification of $a$ into a set of possible modification types $\in m$. An instance of $a$ is then a tuple ${<}governor, relation, dependent{>}$ where one of the arguments is related to a trigger word $p$, and both arguments $\in S$.
\end{defn}

\subsection{Baseline system}
We adapt a span-based approach that has been used previously for the tasks of co-reference resolution \cite{Lee2017}, Semantic Role Labeling \cite{He2018}, and scientific IE \cite{Luan2018a}. The use of span representations as classifier features enables end-to-end learning by propagating information between multiple tasks without increasing the complexity of inference. We train the \textsc{SciIE} system \cite{Luan2018a} on \textsc{FOBIE} to extract spans that constitute trigger words and key-phrases, as well as the binary relations between these spans. 

Figure \ref{fig:example_input} illustrates the input that we provide to \textsc{SciIE}. All tokens are embedded using GloVe \cite{Pennington2014} and ELMo embeddings (original) \cite{Peters2018}. For a single sentence ${D = \{w_1,..., w_n\}}$ all possible spans ${S = \{s_1,...,s_N\}}$ are computed, which are within-sentence word sequences.

The model deals with $\mathcal{O}(n^4)$ possible combinations of spans, where $n$ is the number of words in a sentence. Therefore, pruning is required to make the classification of span-pairs into relation labels tractable at both training and test time \cite{Lee2017, He2018}. First, a score $\phi_{mr}$ of how likely a span is mentioned in a relation is computed. These mention scores enable beam pruning the number of spans considered for relation classification with a variable beam of size $\lambda_{n}$, where $n$ is the number of tokens in the input sentence \cite{Luan2018a}. Second, the maximum width $W$ of spans is limited to reduce the total number of spans. We set $\lambda$ to $.8$ and $W$ to $14$ tokens, the maximum span length in \textsc{FOBIE}.

After pruning, a label $e_{i} \in L_{E}$ is predicted for the remaining spans $s_{i}$ . Here $L_{E}$ is the set of possible span labels, including a non-span class $\epsilon$. For pairs of spans $(s_i, s_j)$ the model predicts which relation $r_{ij}$ $\in L_{R}$ holds between them. The set of possible relation types is $L_{R}$, which includes a non-relation class $\epsilon$. The output consists of labeled spans and relation labels for pairs of spans. For a detailed description of the \textsc{SciIE} system we refer to Luan \textit{et al.} \shortcite{Luan2018a}.

\begin{figure*}[ht]
    \centering
    \includegraphics[width=\textwidth]{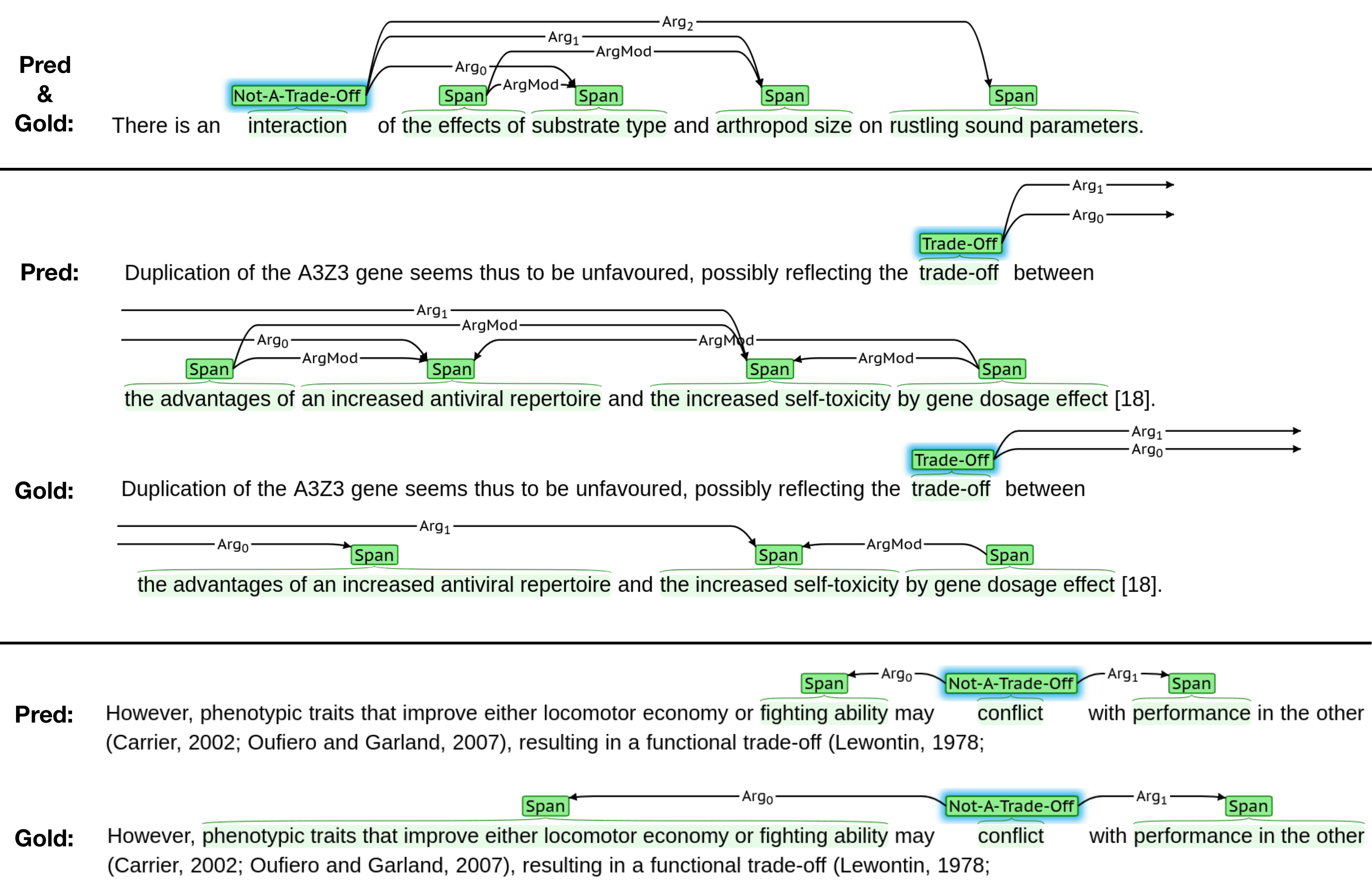}
    \caption{Examples of output from the \textsc{SciIE} system trained on \textsc{FOBIE}, in comparison to gold annotation.}
    \label{fig:example_output}
\end{figure*}{}

\subsection{Narrow IE results}
\label{sec:results}

We evaluate \textsc{SciIE} on two sub-tasks: (1) Argument Recognition, and (2) Relation Extraction. Table~\ref{tab:sciie_vs_our_best} summarizes the results on the sub-tasks of Argument Recognition and RE. With regards to the first sub-task, we train two \textsc{SciIE} models. One model only predicts whether a span is a valid span or not, while a second model predicts whether the span is a trigger word or a key-phrase. For the first sub-task we also report the results of the \textsc{RBS} described in Section \ref{ssec:dataset_creation}. The \textsc{RBS} performs significantly worse; it identifies trigger words exceptionally well (F1=95.89 on test set) but does not correctly recognize many of the remaining key-phrases (F1=22.36 on test set), resulting in a low overall performance.


Figure~\ref{fig:example_output} shows example outputs of the narrow RE model. The predicted relation (\textsc{Not-A-Trade-Off}) and its accompanying structure for the first example are completely correct. Note how the argument modifiers result in a non-projective structure. The second example is more challenging, with a longer range dependency between the trade-off span and the second dependent argument. Our model predicts the correct relation, \textsc{Trade-Off}, but only extracts partial argument spans and essentially fragments them into several modifying argument relations. The third example exhibits a relatively long argument -- which is common in scientific literature -- where only a small part of the span is predicted.

\subsection{Supporting trade-off annotation}
A qualitative analysis confirms the ability of the trained narrow IE system to support a domain expert during trade-off annotation. We predict trade-offs for 523 unlabeled, scientific papers that have been annotated with a trade-off in an ontology of biomimetics \cite{Vincent2014, Vincent2016}. A domain expert compares the trade-offs found in the ontology of biomimetics against the output of the \textsc{SciIE} system, see Table~\ref{tab:julians_bmo}. Narrow IE is found to locate the central \textsc{Trade-Off} relations and arguments for 41.68\% of the total 523 papers. Explicit trade-offs were found in 243 documents. At least one of the extracted \textsc{Trade-Off} relations for each document is identical to the expert annotation in 77.37\% of these documents. For 89.71\% of the 243 documents a trade-off was found to be correct after some interpretation by the expert. Two main types of uninformative trade-offs were found: trade-offs from a cited source and trade-offs between generic terms, e.g., a trade-off between cost and benefit without defining what the cost and benefit are.

\begin{table}[]
    \centering \small
    \begin{tabular}{l|ccc}
  \multicolumn{1}{l}{\textit{Argument Recognition}}
 &  \multicolumn{1}{c}{P}
 & \multicolumn{1}{c}{R} 
 & \multicolumn{1}{c}{F1} \\ \hline
  \textsc{RBS}       &  44.31 & 35.32 & 39.31    \\
  \textsc{SciIE} unlabeled spans & \textbf{84.50}	& 80.40	& 82.40  \\ 
  \textsc{SciIE} labeled spans  & 84.11	& \textbf{80.94}	& \textbf{82.49}   \\
  \multicolumn{1}{l}{\textit{Relation Extraction}}
 &  \multicolumn{1}{c}{P}
 & \multicolumn{1}{c}{R} 
 & \multicolumn{1}{c}{F1} \\ \hline
  \textsc{RBS}       & ---    & ---   & ---       \\
  \textsc{SciIE} unlabeled spans    & 69.60	& \textbf{69.60}	& 69.60    \\ 
   \textsc{SciIE} labeled spans  & \textbf{71.72}	& 68.72	& \textbf{70.19} \\
    \end{tabular}
    \caption{Results on test set for our Rule-Based System (RBS) and \textsc{SciIE} \cite{Luan2018a} w.r.t argument recognition and the combined task of extracting and classifying relations (RE). Providing the model with labels to distinguish trigger words from key-phrases slightly improves performance. 
    }
    \label{tab:sciie_vs_our_best}
\end{table}{}

\begin{table}[]
    \centering \small
    \begin{tabular}{l|l}
        Documents with identified trade-offs    & 243 \\ 
        Exact match                  & 77.37\%     \\
        Match after interpretation   & 89.71\%  \\
         & \\
         Sentences with identified trade-off   & 998 \\
         Exact match                & 68.04\% \\
         Match after interpretation & 84.47\% \\
    \end{tabular}
    \caption{Manual analysis of extractions from 523 scientific documents that were used in the creation of an ontology of biomimetics \cite{Vincent2014,Vincent2016}.}
    \label{tab:julians_bmo}
\end{table}

\begin{table*}[h]
    \small
    \centering 
    \begin{tabular}{lllllll}
        Cluster name                & \textbf{Immunity}              & \textbf{Size}              & \textbf{Locomotion}           \\ \hline
        Top-5 arguments             & immunity               & size              & swimming             \\
                                    & immune function        & number            & sprinting             \\
                                    & the immune system     & volume            & running           \\
                                    & incompetence          & age               & locomotion           \\
                                    & immune response       & time              & diving            \\ \hline
        Top-3 related clusters      & Mating                & Temperature       & Attribute of Animal   \\
                                    & Reproduction          & Sperm Length      & Verbs         \\
                                    & Life History Traits   & Offspring Number  & Capacity/Endurance  
\end{tabular}
    \caption{Examples of clusters found using the K-means algorithm on trade-off arguments from 1279 documents. For the related clusters only \textsc{Trade-Off} relations are taken into account. }
    \label{tab:cluster_insight}
\end{table*}

\section{Semi-Open Relation Extraction} 
\subsection{Task description} \label{ssec:sore_task}

We define the aim of SORE as extracting the relations and concepts in a text that capture the most central information. The application of SORE is especially of interest to scientific IE where OIE systems perform poorly and narrow IE systems are unable to cover the wealth of different relations types. One possible approach is to automatically filter out uninformative and incorrect extractions generated by OIE systems. In this approach, SORE relies on the output of both types of systems, providing a middle ground between precise, narrow IE and unbounded, but unreliable, OIE. The resulting extractions are expected to be useful for human readers, but can also be used to collect data for annotation and training of scientific IE systems.

\subsection{Experimental setup}
We explore SORE on scientific biology texts using the output of the \textsc{SciIE} system trained on \textsc{FOBIE}, predicting trade-offs for the unlabeled 10k open access biology papers (see section \ref{ssec:dataset_creation}). The narrow IE output consists of 2,216 trade-offs found in 1,279 documents. We pre-process arguments by appending their modifier, removing stop words, and embedding the remaining sequences using ELMo (PubMed)\footnote{https://allennlp.org/elmo}. We use the K-means algorithm to compute clusters on the IDF-weighted average of the resulting argument representations. 
A domain expert inspected the centroids qualitatively. Table~\ref{tab:cluster_insight} provides insight into some of the resulting argument clusters and their interrelations. The exact number of clusters does not seem to greatly affect SORE. For the given narrow IE output $\pm$50 clusters seems to provide a good balance between generic and more fine-grained topics.
The IDF weights are computed over the subword units found in the dataset; we use SentencePiece\footnote{https://github.com/google/sentencepiece} with a vocabulary of 16K. We then run OpenIE~5, a state-of-the-art OIE system \cite{Saha2018,Saha2017,Pal2016,Christensen2011}, on the same 1,279 documents that were found to contain one or more \textsc{Trade-Off} relations. 

We retain only OIE extractions that contain one or more arguments that are classified into the same cluster as the \textsc{Trade-Off} arguments found in that text. Furthermore, we omit OIE arguments that belong to noisy clusters containing mostly math symbols or long nested phrases. We compute a simple IDF-weighted cosine similarity \cite{Galarraga} between the vector representations of the remaining OIE and trade-off arguments.

\subsection{Qualitative analysis of SORE output} \label{sec:uninformative}
We notice a striking drop in the number of irrelevant and noisy OIE arguments that remain after applying SORE. The total amount of OIE extractions reduces from 401k before filtering to 140k (34.95\%) after filtering. As a result, the number of OIE extractions per document reduces from 314 to 110. The unfiltered OIE extractions are found in 170k sentences, of which 67k (39.55\%) are retained after applying SORE. 

To test our hypothesis that SORE can reduce the number of uninformative extractions, without limiting RE to a narrow set of relations, we randomly select representative samples of unfiltered and filtered OIE extractions (400 each). A domain expert manually annotated whether each extraction or sentence was thought to be informative, e.g., provides relevant information to understanding a biological text. As an example, consider the sentence ``\textit{We have used this approach in a previous study to investigate the molecular factors governing the altered liver regeneration dynamics caused by ablation of the gene adiponectin (Adn)''} \cite{Cook2015}. OIE extractions such as `\textit{(We, have used, this [...] study)}' are considered uninformative, in contrast to `\textit{(the molecular [...] dynamics, caused by, ablation [...] adiponectin)}'.

Many OIE extractions are found to be poorly structured. Like Groth \textit{et al.} \shortcite{Groth2018} we relax the requirement of extractions being well-formed, e.g., we consider extractions that incorrectly identify the boundaries of one or more arguments as possibly capturing relevant information. Different from their evaluation on correctness, we evaluate whether an extraction captures information that is relevant to understanding a text. As a result, we consider poorly structured OIE extractions that contain relevant information to be informative, e.g.:
\begin{itemize}
    \item \textit{('the resumption of respiration', ' can lead to an increase of superoxide anions in the cytosol perhaps driving', ' increased elevation of Cu-ZnSOD')}.
    \item \textit{('transcriptional coregulation amongst many genes', ' will give', ' rise to indirect interaction effects in mRNA expression data')}.
\end{itemize}

The annotation relies on the correctness of the information captured by OIE extractions and whether this information is useful to a reader. However, this does not imply informative extractions are relevant to the central theme of the text captured in a trade-off. We consider OIE extractions uninformative if the extraction:
\begin{itemize}
    \item contains an uninformative argument class, e.g., \textit{('Miller et al . , 2012', ' to minimize', ' their swimming effort')}.
    \item contains incomplete arguments, e.g., \textit{ ('the RDME requirement', ' reactions', ' only fire')}.
    \item is non-sensible, e.g., \textit{('P. magellanicus', ' would have resulted', ' in a 1.6-fold higher Vmax for the scallop muscle')}.
    \item is unlikely to help understand a text, e.g., \textit{('DeepBind', ' was trained', ' on data from RNAcompete , CLIP - RIP - seq [ 10')} and \textit{('microlepidopteran superfamilies', ' are heavily entombed', ' L:in amber')}.
\end{itemize}

We also randomly select representative samples from the 170k unfiltered and 67k filtered sentences from which the OIE extractions are sourced. The reason is that erroneous OIE extractions, e.g., not well-formed tuples, can guide a reader to informative passages in a text. We see similar errors as described by Schneider \textit{et al.} \shortcite{Schneider2017} and Groth \textit{et al.} \shortcite{Groth2018}, e.g., long sentences lead to incorrect extractions and errors in argument boundaries. To illustrate the complexity of sentences that an OIE system encounters in scientific texts, consider the following examples:
\begin{itemize}
    \item the arity of relations can be high, e.g., (49 tokens) ``\textit{A large genome size tends to correlate with delayed mitotic and meiotic division [6–8] decreased plant invasiveness of disturbed sites [9] lower maximum photosynthetic rates in plants [2] and lower metabolic rates in mammals [10] and birds [11, 12].}'' \cite{Warringer2006}.
    \item many phrases are nested and express non-verbal relations, e.g., (45 tokens) ``\textit{However, for arboreal animals that regularly jump between branches (often when elevated quite high above the ground), jumping \textbf{accurately} (\textbf{which we define as the ability to land close to the intended target}) may also be important to fitness.}'' \cite{Kuo2011}.
\end{itemize}

Table~\ref{tab:informativeness} provides an overview of the annotation results. Filtering is found to increase the informativeness of both OIE extractions ($\chi^2$=6.39, p$<$.025) and sentences ($\chi^2$=11.75, p$<$.01). The percentage of informative OIE extractions increases by 5.75\% and of the percentage of informative sentences by 8.25\%.  A second domain expert annotated 25\% of each set (400 total), the inter-annotator agreement Cohen \textit{k} was found to be 0.84.

\begin{table}[]
    \centering \small
    \begin{tabular}{@{$\,$}llcc@{$\,$}}
          & \footnotesize\textbf{Filtering} &\footnotesize\textbf{\# Nr.}&\footnotesize\textbf{\% Informative}\\ \hline \rule{0pt}{2ex} 
        \multirow{2}{*}{\textit{OIE extractions }} & before  & 401,588   & 29.25\% \\ %
                                & after          &  140,357   & 35.00\% \\ \cline{2-4}\rule{0pt}{2ex} 
        \multirow{2}{*}{\textit{Sentences}}  & before   & 170,551  & 36.50\% \\ %
                                & after      & 67,460    & 44.75\%
        
    \end{tabular}
        

    \caption{Total number of OIE extractions before and after filtering, as well as the sentences that these extractions were found in. The \% informative denotes the percentage of extractions and sentences annotated as informative by a domain expert, based on 400 randomly sampled instances from each group (95\% confidence interval, margin of error 5\%).}
    \label{tab:informativeness}
\end{table}

\subsection{Results}
Manual inspection of the retained OIE extractions shows that many relevant extractions are retained, e.g., see Table~\ref{tab:SORE_example}. These extractions are useful to a reader in determining whether a document is worth reading in full, and can be used to identify informative sections in a text. The presented approach to SORE shows promising results w.r.t. automatically filtering out a large proportion of irrelevant, incorrect, or uninformative OIE extractions. Considering the poor quality of OIE extractions, however, we propose presenting a reader with the sentences that entail the filtered OIE extractions. Furthermore, SORE provides a method to collect data for annotation and training of scientific OIE systems.


\begin{table}[ht]
\small \centering
\begin{tabular}{p{.94\linewidth}}
\textbf{\textsc{Trade-off} relations} \\ \hline
\begin{tabular}{p{.42\linewidth}p{.45\linewidth}}
    \textit{Trade-off arguments}  & \textit{Argument modifiers  }                      \\ \hline \rowcolor{lightgray}
    sleep               & \\ \rowcolor{white}
    cognitive abilities & \\ \rowcolor{lightgray}
    energy conservation & \\ \rowcolor{white}
    memory retention    & (the keeping of memory over prolonged periods of time) \\ \rowcolor{lightgray}
    memory consolidation & {(in bats) \newline (without a food reward) \newline (shift from short- to long-term memory) \newline (using torpor) } \\ 
\end{tabular}


\rule{0pt}{2ex} \textbf{Examples filtered OIE extractions} \\ 
\hline 

\begin{tabular}{p{.93\linewidth}}
    \textit{(A memory; is normally formed; after repeated learning events; sleep enhances this process)}\\  \rowcolor{lightgray}
    \textit{(learning; is associated; with a food reward)} \\ \rowcolor{white}
    \textit{(Sleep deprivation; has; negative effects on both memory consolidation)}   \\ \rowcolor{lightgray}
    \textit{(torpor; has; a negative influence on memory consolidation)}  \\ \rowcolor{white}
    \textit{(digestion; prevents; the bats; from falling into torpor quickly)}\\ \rowcolor{lightgray}
    \textit{(torpor ; indeed affects ; learning abilities)} \\ \hline
\end{tabular}
\end{tabular}
\caption{SORE extractions from a scientific biology text \cite{Ruczyski2014DoLearning}. The \textsc{Trade-Off} relations are extracted by a narrow IE system trained on \textsc{FOBIE}. These relations capture the central theme and concepts of the text, and are used to filter the extractions that an OIE system outputs for the same document. The resulting extractions can support discerning the relevance of scientific documents.} \label{tab:SORE_example}
\end{table} 

\section{Conclusions}
We introduce the task of Semi-Open Relation Extraction (SORE) on scientific texts and the Focused Open Biological Information Extraction (FOBIE) dataset. We adapt off-the-shelf IE systems to show that SORE is feasible, and that our approach is worth improving upon -- both in terms of performance, as well as reducing the system’s complexity. A strong scientific IE system is used as a baseline, and its output is used to filter the relations found by a state-of-the-art OIE system. 


OIE from scientific text is a hard task. The large number of errors that we find in OIE extractions from scientific texts render them near-useless to downstream computing tasks. A human reader may, nevertheless, find many incorrect extractions informative. An issue for humans is the sheer amount of OIE extractions and the high proportion of uninformative extractions. We show that our approach to SORE reduces the number of OIE extractions by 65\%, while increasing the relative amount of informative extractions by 5.75\%. As a result, SORE improves the ability for a reader to quickly skim through the remaining extractions, or sentences that they are sourced from, and analyze how central concepts are related in a scientific text. 

The presented approach is currently limited to the domain of biology and the use of trade-off relations, but we expect that central relations can be identified for other scientific domains that enable SORE. We show that creating a dataset for narrow RE can be done relatively cheaply by re-annotating the output of a simple RBS. Similarly, SORE may aid the collection of a dataset for scientific OIE.

\section*{Acknowledgments}
The authors would like to gratefully acknowledge the financial support of the Engineering and Physical Sciences Research Council (EPSRC) Centre for Doctoral Training in Embedded Intelligence under grant reference EP/L014998/1 and the EPSRC Innovation Placement fund. We also thank Ben Trevett for proof-reading the document.

\bibliography{Mendeley}
\bibliographystyle{acl_natbib}

\end{document}